**A deep learning pipeline for PAM50 subtype classification using histopathology images and multi-objective patch selection**

Arezoo Borji [1, 2, 3], Gernot Kronreif [1], Bernhard Angermayr[2, 4], Francisco Mario Calisto[5], Ali Abbasian Ardakani [2], Wolfgang Birkfellner [3], Inna Servetnyk[2], Yinyin Yuan[6], Sepideh Hatamikia [2, 1, 3*]

1 Austrian Center for Medical Innovation and Technology, Wiener Neustadt, Austria
2 Department of Medicine, Faculty of Medicine and Dentistry, Danube Private University, Krems an der Donau, Austria.
3 Department of Medical Physics and Biomedical Engineering, Medical University of Vienna, Vienna, Austria
4 Patho im Zentrum, Saint Pölten, Austria
5 Institute for Systems and Robotics (ISR-Lisboa/LARSyS), Instituto Superior Técnico, University of Lisbon, 1049-001, Lisbon, Portugal
6 The University of Texas MD Anderson Cancer Center, Houston, TX, United States

## Abstract

Breast cancer is a highly heterogeneous disease with diverse molecular profiles. Recent AI models have attempted to predict PAM50 subtype, as a standard for classifying breast cancer into intrinsic subtypes, from histopathology images, most depend on random patch sampling that introduces redundancy and restricts the model performance.

In this study, we introduce a novel optimization-driven deep learning framework that aims to reduce reliance on costly molecular assays by directly predicting PAM50 subtypes from H&E-stained whole-slide images (WSIs). Our method jointly optimizes patch informativeness, spatial diversity, uncertainty, and patch count by combining the non-dominated sorting genetic algorithm II (NSGA-II) with Monte Carlo dropout-based uncertainty estimation. The proposed method can identify a small but highly informative patch subset for classification. We used a ResNet18 backbone for feature extraction and a fully connected head for classification. For evaluation, we used the internal TCGA-BRCA dataset as the training cohort and the external CPTAC-BRCA dataset as the test cohort. On the internal dataset, an F1-score of 0.8964 and an AUC of 0.9865 using 627 WSIs from the TCGA-BRCA cohort were achieved. The performance of the proposed approach on the external validation dataset showed an F1-score of 0.7995 and an AUC of 0.9523.

These findings indicate that the proposed optimization-guided, uncertainty-aware patch selection can achieve high performance and improve the computational efficiency of histopathology-based PAM50 classification compared to existing methods, suggesting a scalable imaging-based replacement that has the potential to support clinical decision-making.

## 1. Introduction

Breast cancer is one of the most common and challenging cancers in women because of its pronounced molecular heterogeneity and different therapeutic responses [1]. To handle this complexity, molecular classification tools are utilized which can enhance individualized treatment plans. The PAM50 gene signature, which examines the activity of 50 specific genes [2], has become an important way to categorize breast cancer into different types, such as Luminal A, Luminal B, HER2-enriched, Basal-like, and Normal-like[3]. Even though PAM50 subtyping is helpful for treatment, it usually depends on expensive tests like RNA sequencing [4], which are often not practical in many everyday medical situations, especially in countries with limited resources [5].

Histopathological examinations of breast biopsies are known as the gold standard for cancer diagnosis and grading [6], and they can be used to predict molecular subtypes [7]. AI-driven approaches could provide a cost-effective solution, enabling molecular-level insights while reducing the need for additional laboratory assays ([6],[8]). Deep learning models, particularly convolutional neural networks (CNNs), have demonstrated significant success in histopathology image analysis [9], achieving reasonable performance in cancer diagnosis [10], tumor grading, and biomarker prediction [11]. Several studies have explored the prediction of PAM50 breast cancer subtypes using different data modalities, including gene expression, clinical variables, radiomics, and histopathology [12-18].

Among these, histopathology images are particularly advantageous, as H&E-stained WSIs are routinely available in clinical workflows and capture rich morphology linked to molecular characteristics, making them a practical and scalable choice for subtype prediction [15].

In recent years, different studies have applied CNNs to histopathology WSIs for PAM50 breast cancer subtype prediction. For example, Ektefaie et al. [19] developed weakly supervised models on H&E WSIs to predict receptor status, PAM50 subtypes, and TP53 mutation with an accuracy of 65.4% for PAM50 with validation on independent cohorts. Phan et al. [20] introduced a two-step transfer-learning pipeline, first pretraining on an in-house pathology slides and then fine-tuning on external TCGA-BRCA cohort . Using ResNet101 on unannotated H&E WSIs, they achieved 0.78 patch-level accuracy and 0.913 slide-level accuracy. Tafavoghi et al. [21] proposed a two-stage pipeline that first separated tumor from non-tumor patches, followed by PAM50 classification using ResNet18 features combined with XGBoost. Their method achievied a macro-F1 score of 0.73 on a 221-case hold-out set. Complementing subtype classification, Kurian et al. [22] trained a deep CNN on TCGA-BRCA H&E WSIs to quantify intratumor heterogeneity (ITH) in Luminal A tumors. Their model revealed strong associations between subtype admixture and molecular alterations (e.g., TP53, PIK3CA) as well as significantly worse progression-free survival in more admixed cases.

Jaber et al. [14] developed a multiscale patch-based pipeline on H&E-stained WSIs using Inception-v3 to classify breast cancer into four PAM50 subtypes. Their method achieved an accuracy of 65.92%. Similarly, Liu et al. [15] proposed a multi-instance learning framework (DPMIL) on H&E WSIs using LOF-based patch filtering, with their weighted fusion model reaching an F1-score of 69 % and an accuracy of 74.85%.

Most existing pathology studies utilized rule-based sampling patch extraction rather than optimization strategies. Common approaches include uniform grid-based tiling [23], random sampling [24], naive grid [25], and fixed random sampling [26]. Although these strategies simplify the preprocessing pipeline, they often result in large numbers of redundant, noisy, or non-informative patches, increasing computational burden and limiting model discriminative power.

Despite recent advances in CNN-based histopathology methods for PAM50 prediction, important methodological gaps remain. Current pipelines predominantly rely on heuristic or rule-based patch selection strategies [14, 15, 19–26]. In addition, most published models provide only point estimates of subtype probability and do not incorporate formal modeling of predictive uncertainty [10-16]. Furthermore, generalizability remains limited because many reports evaluate performance solely on internal splits and external validation across cohorts is limited [14-15].

Patch selection is an important step in histopathology image analysis because WSIs contain many patches, many of which are histologically redundant or have little diagnostic value [27]. From a clinical standpoint, only a subset of regions accurately reflects the underlying tumor biology and guides diagnostic and prognostic decisions. By systematically optimizing patch selection, we focus the model's attention on diagnostically relevant and morphologically diverse tissue regions that pathologists typically rely on for subtype assessment, thus improving predictive accuracy. Training models on unfiltered patches introduces noise, increases computational burden [28], and can obscure clinically significant patterns [29].

Recently, few studies have worked on optimization-based patch selection to overcome the limitations of rule-based sampling in WSI analysis and capture more clinically meaningful tissue regions [30]. Zheng et al. (2024) [31] introduced a dynamic policy-driven adaptive multi-instance learning (PAMIL) model using the WSI (H&E) framework to integrate reinforcement learning (RL) with dynamic instance sampling. Their method selected the most informative patches during training, which improved feature aggregation and decision-making. This strategy yielded a 3.8% improvement on the CAMELYON16 and 4.4% on the TCGA-lung datasets compared to conventional approaches. Similarly, Raza et al. [32] proposed a dual-attention reinforcement learning model using WSI to optimize patch selection through a learned reward function. Their method could effectively prioritize diagnostically relevant regions and achieved better classification performance. In parallel, Cauteruccio et al. [33] proposed AgentViT, a reinforcement learning–based patch selection framework for vision transformers, where the agent learns to balance classification performance and the number of selected patches in WSI data. Although this approach demonstrated the effectiveness of optimization for patch selection, it primarily targeted computational efficiency in general computer-vision benchmarks rather than clinical histopathology tasks. Liu et al. [15] applied a local outlier factor (LOF) algorithm following a dual-network training method to filter noise patches and identify discriminative regions. While this approach improved model performance, it relies on local density estimation and rule-based filtering rather than optimization formulation. Despite these advances, current optimization-based patch selection methods remain limited in scope; they primarily rely on single reward functions and do not jointly optimize multiple complementary objectives such as patch informativeness, uncertainty, diversity, and patch count.

Predictive uncertainty is an important aspect of developing reliable and clinical AI systems [34]. In clinical decision-support settings, uncertainty estimation detects unreliable predictions caused by tissue heterogeneity, imaging artifacts, and diagnostically ambiguous regions [35]. Models that quantify uncertainty can better manage diagnostic risk, allow for selective expert review, and support safer and more transparent clinical deployment in pathology.

Although uncertainty estimation has not been explicitly addressed in previous PAM50 classification studies [10–16], it has been recognized as both clinically and technically important in the broader medical AI literature. The generative segmentation approach proposed by Kohl et al. [36] addresses intrinsic image ambiguity through the modeling of probable outputs; however, its segmentation-specific design and complexity hinder its direct application to classification tasks. To identify predictions that are inaccurate or out of distribution, Hendrycks and Gimpel [37] employed confidence-based uncertainty estimation; nevertheless, their heuristic technique lacks adequate calibration. One of the most influential approaches in this area is Monte Carlo Dropout, which approximates Bayesian inference by applying dropout during test time, enabling estimation of epistemic uncertainty without modifying the underlying network architecture [38]. This method is computationally efficient compared to classical Bayesian approaches but still requires multiple stochastic forward passes, which may increase inference time and can produce overconfident estimates on out-of-distribution data.

Furthermore, nearly all the studies were single-center investigations lacking external validation [10, 11, 14–16]. Ektefaie et al. [19] studied an independent cohort and reported a modest PAM50 classification accuracy of 65.4%. In contrast, Phan et al. [20] achieved the highest reported performance, with a slide-level accuracy of 0.913, despite the lack of documented external testing results.

To address these limitations, our proposed framework formulates a patch selection model using a multi-objective optimization problem, which balances informativeness, diversity, and patch count using NSGA-II, thereby enabling a more structured and data-driven selection process. Additionally, we incorporated Monte Carlo (MC) Dropout into our pipeline to quantify patch-level predictive uncertainty and use it as an optimization signal in NSGA-II patch selection. Our approach specifically includes prediction uncertainty as an optimization goal in NSGA-II. By doing this, the model can optimize variety and patch budget while focusing on clinically significant ambiguous locations since uncertainty actively directs patch selection. Furthermore, we validate our proposed model on an external dataset, CPTAC-BRCA, to show the generalization of our model.

In summary, our main innovation lies in a two-stage patch selection strategy that combines uncertainty-guided filtering with multi-objective optimization using NSGA-II. In the first stage, patches with low predictive uncertainty are prioritized, and unreliable regions are filtered out. In the second stage, NSGA-II jointly optimizes informativeness, morphological diversity, and compactness while minimizing residual uncertainty, thereby directing the model's focus toward the most pertinent tissue areas. Our results indicate that our method can enhance both accuracy and generalizability compared to earlier studies, showing strong performance on both the internal TCGA-BRCA and external CPTAC-BRCA test datasets.

## 3. Methodology

### 3.1. Dataset description

This study uses 627 high-resolution WSIs from the TCGA-BRCA open access dataset [21]. Each slide is H&E-stained and represents one of the four intrinsic PAM50 molecular subtypes: Luminal A (339 slides), Luminal B (131 slides), HER2-enriched (51 slides), and Basal-like (106 slides). To ensure morphological fidelity and histopathological integrity, only slides from formalin-fixed paraffin-embedded (FFPE) tissue specimens are used in the analysis to preserve structural features [39]. To evaluate the generalizability of our model, we tested its performance on an external dataset , the CPTAC-BRCA open access dataset [40], which contains 122 H&E-stained FFPE WSIs with subtype annotations, including Luminal A (51 slides), Luminal B (28 slides), HER2-enriched (12 slides), and Basal-like (31 slides).

For TCGA-BRCA, we performed a slide-level split at the patient-level to prevent data leakage: 80% of patients were used for training and 20% for validation. When multiple WSIs were available for a given patient, all slides were assigned exclusively to the same split. For CPTAC-BRCA, all 122 WSIs were reserved for external testing.

### 3.2. Subtype annotation using IHC surrogates

In clinical documents where transcriptomic profiling is not available, immunohistochemistry (IHC) is often utilized as an alternative for molecular subtyping of breast cancer, including PAM50-aligned intrinsic subtypes. The classification based on IHC depends on the expression levels of four key biomarkers, estrogen receptor (ER), progesterone receptor (PR), human epidermal growth factor receptor 2 (HER2), and Ki-67 commonly evaluated in clinical practice to inform treatment choices. In this investigation, metadata related to receptor status (ER, PR, HER2, and Ki-67) was extracted from the clinical records of TCGA-BRCA and used to categorize molecular subtypes when transcriptomic PAM50 annotations were not obtainable. Subtypes were classified according to established clinical guidelines: Luminal A (ER+, PR+, HER2−, low Ki-67), Luminal B (ER+, HER2±, high Ki-67), HER2-enriched (HER2+, ER−, PR−, high Ki-67), and Basal-like/Triple-negative (ER−, PR−, HER2−, high Ki-67). These classifications serve as a clinically relevant approximation of intrinsic PAM50 subtypes and are frequently employed in routine oncology practices as well as retrospective analyses. For the TCGA-BRCA cohort (N = 627 WSIs), PAM50 subtype labels were obtained from transcriptomic PAM50 results for 426 WSIs, while the remaining 201 WSIs were classified using IHC surrogate markers. Although this method is clinically significant, the subtype assignment based on IHC provides an estimated representation of transcriptomic PAM50 profiling and introduces additional label variability, particularly at subtype distinctions like Luminal A versus Luminal B. Thus, the findings of this study should be regarded as predicting IHC-aligned intrinsic subtypes, which may not always perfectly align with gene expression–based PAM50 classifications.

For the TCGA-BRCA cohort (N = 627 WSIs), PAM50 subtype labels were derived from transcriptomic PAM50 calls for 426 WSIs and from IHC surrogate markers for the remaining 201 WSIs. In the absence of transcriptomic profiling, molecular subtypes were assigned using ER, PR, HER2, and Ki-67 status according to established clinical criteria. While these IHC-based definitions are widely used in routine clinical practice, they introduce additional label noise compared to transcriptomic PAM50 calls; therefore, the reported results should be interpreted as predicting IHC-aligned intrinsic subtypes, which may not perfectly coincide with gene expression–based PAM50 classifications in all cases.

### 3.3. Patch-wise extraction and preprocessing of WSIs

The WSIs underwent processing through a standardized pipeline to guarantee high-quality, consistent inputs across the domain for subsequent deep learning analysis. All WSIs were accessed at Level 0, which corresponds to the highest spatial resolution of the digital scans (40× magnification, 0.25 μm/pixel for TCGA-BRCA), thus retaining essential fine-grained morphological details necessary for distinguishing breast cancer subtypes. Each WSI, having a width $W$ and height $H$, was divided into non-overlapping patches, each measuring 512 × 512 pixels, as depicted in Figure 1. Only patches fully contained within the image boundaries were retained, yielding a total of extractable patches equal to:

$$N=\left\lfloor\frac{w}{512}\right\rfloor\times\left\lfloor\frac{H}{512}\right\rfloor \qquad (1)$$

All patches were then adjusted to a size of 224 × 224 pixels to align with the input specifications of pretrained CNN architectures. To eliminate regions dominated by background, a simple tissue detection method was utilized. Each RGB patch was transformed into grayscale by employing a luminance-preserving weighted combination:

$$G=0.2989\cdot R+0.5870\cdot G+0.1140\cdot B \qquad (2)$$

The tissue fraction $M$ within each patch was computed as the proportion of pixels with grayscale intensity below 200, indicating non-background tissue:

$$M=\frac{1}{n}\sum_{i=1}^{n}(G_i < 200) \quad , n=512\times512=262{,}144 \qquad (3)$$

Only patches with $M\geq 0.2$ were selected for subsequent analysis. This threshold aligns with earlier research in digital pathology, which typically employs tissue-content cutoffs of 10% to 30% to eliminate regions dominated by background. A practical review validated that a cutoff of 20% successfully eliminated void spaces while maintaining tissue structures that are important for diagnosis.

To identify and discard visually compromised patches, blur was quantified using the variance of the Laplacian operator:

$$Var_L=\mathrm{Var}(\nabla^2 I) \qquad (4)$$

Patches with $\mathrm{Var}_L < 100$were considered out-of-focus and removed, consistent with commonly used quality-control thresholds in WSI preprocessing. Additional heuristic filters are used to remove patches dominated by slide borders.

To reduce the inter-laboratory color variability inherent in H&E-stained slides, stain normalization was performed using the Macenko method, which standardizes hematoxylin and eosin stain vectors across WSIs. After stain normalization and resizing, patches were normalized with ImageNet statistics to align their intensity distributions with those used during CNN pretraining.

$$R_{norm}=\frac{R-0.485}{0.229},\quad G_{norm}=\frac{G-0.456}{0.224},\quad B_{norm}=\frac{B-0.406}{0.225} \qquad (5)$$

This normalization improves training stability and transfer learning efficiency by matching the feature distribution expected by pretrained CNN backbones.

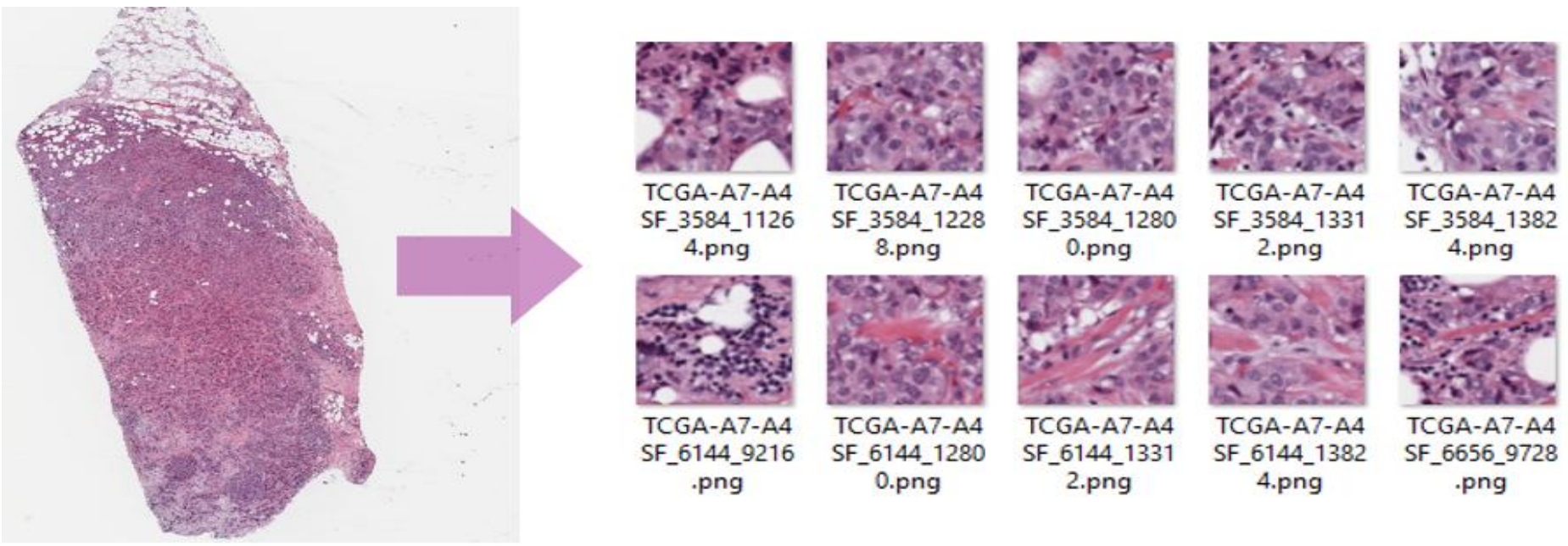

`**Fig.1.** A HER2-enriched breast cancer WSI divided into patches for subtype classification.

### 3.4. Feature extraction using ResNet-18

Each input patch $X_i$ was passed through a ResNet-18 network pretrained on ImageNet. The final classification head was removed, and all convolutional layers were kept frozen, using the 512-dimensional penultimate layer as a fixed embedding.

$$Z_i = f_{\text{ResNet}}(X_i), Z_i \in \mathbb{R}^{512} \tag{7}$$

These embeddings capture high-level histomorphology patterns, including texture, tissue architecture, and cellular organization, and were reused consistently across all downstream stages of the pipeline, including Monte Carlo dropout–based uncertainty estimation and NSGA-II–based patch selection.

### 3.5. Uncertainty estimation via Monte Carlo dropout

To quantify predictive uncertainty, we employed MC dropout, which approximates Bayesian inference by activating dropout at inference time [48]. This setup enables estimation of epistemic uncertainty, which reflects uncertainty in model parameters arising from limited data [49].

During inference, each patch $X_i$ was passed through the classification head $T = 20$ times with dropout enabled at a rate of 0.5, producing $T$ stochastic softmax probability vectors,

$$\{P_i^1, P_i^2, \ldots, P_i^c\}, \quad P_i^{(t)} \in R^c \tag{8}$$

where $C$ denotes the number of target classes and $P_i^{(t)}$ corresponds to the softmax output from the $t$-th stochastic forward pass.

For each class $c$, the variance of predicted probabilities across the $T$ passes was computed as

$$\text{Var}(\, P_i^{(c)}) = \frac{1}{T-1} \sum_{t=1}^{T} (P_{i,c}^{(t)} - \underline{P}_{i,c})^2 \,, \quad \text{where} \;\; \underline{P}_{i,c} = \frac{1}{T} \sum_{t=1}^{T} P_{i,c}^{(t)} \tag{9}$$

where low variance indicates consistent predictions and high variance reflects increased uncertainty.

The overall uncertainty associated with patch $X_i$ was defined as the mean variance across all classes,

$$\text{Uncertainty}\,(X_i\,) = \frac{1}{C} \sum_{c=1}^{C} Var(\, P_i^{(c)}) \tag{10}$$

This scalar uncertainty score was subsequently used to guide patch selection, with higher-uncertainty patches deprioritized to reduce the influence of unreliable or ambiguous regions in downstream analysis. The choice of $T = 20$ represents a practical trade-off between the stability of uncertainty estimates and computational efficiency; in practice, MC dropout increased inference time linearly with $T$, while providing sufficiently stable uncertainty estimates for patch filtering.

## 3.6. Patch selection using predictive uncertainty and multi-objective optimization

WSIs mostly yield thousands of tissue patches, many of which are visually redundant, diagnostically ambiguous, or computationally inefficient to process. To address this, we employed a two-stage patch selection framework that combines uncertainty-guided filtering with evolutionary multi-objective optimization, enabling the extraction of compact, reliable, and morphologically diverse patch subsets for downstream classification.

### 3.6.1. Uncertainty-guided filtering

As an initial filtering step, epistemic uncertainty estimated via Monte Carlo dropout (Section 3.5) was used to remove unreliable patches. Each patch $X_i$was evaluated over $T = 20$stochastic forward passes, and a scalar uncertainty score was computed as the mean class-wise variance of softmax probabilities,

$$\text{Uncertainty}(X_i) = \frac{1}{C}\sum_{c=1}^{C} \text{Var}\,(P_{i,c}). \qquad (11)$$

Patches exhibiting high uncertainty were discarded at this stage, ensuring that only high-confidence tissue regions were retained for subsequent optimization. This step reduced noise and stabilized the downstream multi-objective optimization process.

### 3.6.2. Multi-objective patch selection with NSGA-II

Following uncertainty filtering, patch selection was formulated as a multi-objective optimization problem and solved using the Non-dominated Sorting Genetic Algorithm II (NSGA-II) [51]. Each candidate solution corresponds to a subset of patches represented as a binary vector $s \in \{0,1\}^N$, where $N$denotes the number of available patches after filtering and $s_i = 1$indicates inclusion of patch $i$.

The optimization jointly maximizes four complementary objectives:

- **Diversity**: maximize morphological heterogeneity among selected patches,

$$\phi_1(s) = \frac{2}{|\, s \,|\,(|\, s \,| - 1)} \sum_{i<j,\, i,j\in s} \left(1 - \frac{z_i^{\top} z_j}{\|\, z_i \,\|_2 \|\, z_j \,\|_2}\right), \qquad (12)$$

where $z_i \in \mathbb{R}^{512}$denotes the fixed ResNet-18 embedding of patch $i$.

- **Informativeness**: maximize feature saliency,

$$\phi_2(s) = \frac{1}{|s|}\sum_{i\in s} \| z_i \|_2, \tag{13}$$

- **Compactness**: minimize subset size,

$$\phi_3(s) = -|s|, \tag{14}$$

- **Reliability**: minimize predictive uncertainty,

$$\phi_4(s) = -\frac{1}{|s|}\sum_{i\in s} u_i , \tag{15}$$

where $u_i$ is the Monte Carlo dropout–based uncertainty score of patch $i$.

NSGA-II was executed independently for each WSI in an offline manner prior to model training, using a population size of 50 individuals over 50 generations. Binary tournament selection (tournament size = 2) simulated binary crossover (probability = 0.9), and bit-flip mutation (probability = 0.1) were employed. Candidate solutions were ranked via non-dominated sorting, with crowding distance used to preserve diversity within the Pareto front.

On the TCGA-BRCA cohort, each WSI yielded a median of approximately 10,000 tissue patches after preprocessing. NSGA-II reduced this to a median of approximately 500 patches per slide, corresponding to an approximate 95% reduction in patch count. The optimization was performed offline on a per-slide basis prior to training and required approximately 3 minutes per WSI on a single GPU. This process produced compact, diverse, and uncertainty-aware patch subsets that were subsequently used for subtype classification.

Importantly, the same preprocessing patch selection and processing pipeline was applied to the CPTAC-BRCA cohort, ensuring consistent domain alignment and enabling a fair assessment of model generalizability across datasets.

Algorithm 1: Pseudo code for the patch selection using NSGA-II

**Input:**

$\mathscr{E} = \{X_1, ..., X_N\}$ → Extracted patches from a WSI , f_θ → Trained model with MC Dropout, $T \in \mathbb{N}$ → Number of stochastic forward passes, $M \in \mathbb{N}$ → Population size, $G \in \mathbb{N}$ → Number of generations, $p_c \in [0,1]$ → Crossover probability, $p_m \in [0,1]$ → Mutation probability, $\tau \in \mathbb{R}^{+}$ → Uncertainty threshold

**Output:**

$\mathscr{F}_1$ → Final Pareto front (non-dominated patch subsets)

**Computing epistemic uncertainty via MC Dropout**

**For** each $X_i \in \mathscr{E}$:

$P_i \leftarrow \left\{f_\theta^{(t)}(x_i)\right\}_{t=1}^{T}$ , $u_i \leftarrow \frac{1}{C}\sum_{c=1}^{C} Var(\left\{P_{i,c}^{(t)}\right\}_{t=1}^{T})$

**Filter patches by uncertainty:**

$\mathscr{F} \leftarrow \{X_i \in \mathscr{E} \mid u_i \leq \tau\}$, re-index $F=\{x_1''$,…,$x_{N'}'\}$, $N' \leftarrow |\mathscr{F}|$

**For** each $Xi' \in \mathscr{F}( i = 1, ..., N')$:

$z_i \in R^d \leftarrow$ feature embedding

**Initialize NSGA-II population:**

$P_0 = \left\{s_j \in \{0,1\}^{N'}\right\}_{j=1}^{M}$

**For** generation g = 1 to G:

**For** each $s \in P_{\{g-1\}}$:

$S \leftarrow \{i \in \{1,…,N'\} | s_i = 1\}$

$\varphi_1(S) \leftarrow \frac{2}{|S|(|S|-1)}\sum_{i<j,i}^{,j \in s}(1 - \frac{z_i.z_j}{\|z_i\|2\|z_j\|2})$ → Diversity , $\varphi_2(S) \leftarrow \frac{1}{|s|}\Sigma_{i\in}\|z_i\|2$ → Informativeness , $\varphi_3(S) \leftarrow -|S|$ → Compactness,

$\varphi_4(S) \leftarrow -\frac{1}{|S|}\Sigma_{i\in}u_i$

→ Reliability

Perform non-dominated sorting on $P_{g-1}$

Compute crowding distance for each individual

Select parents via binary tournament (rank + distance)

**Generate offspring via:**

**Crossover** ($s_1$, $s_2$) with probability $P_c$

**Mutation** (bit-flip) with probability $P_m$

$P_{temp} \leftarrow P_{g-1} \cup$ Offspring

$P_G \leftarrow$ Select top M individuals from $P_{temp}$

**Return** final Pareto front:

$\mathscr{F}_1 \leftarrow$ non-dominated set from $P_G$

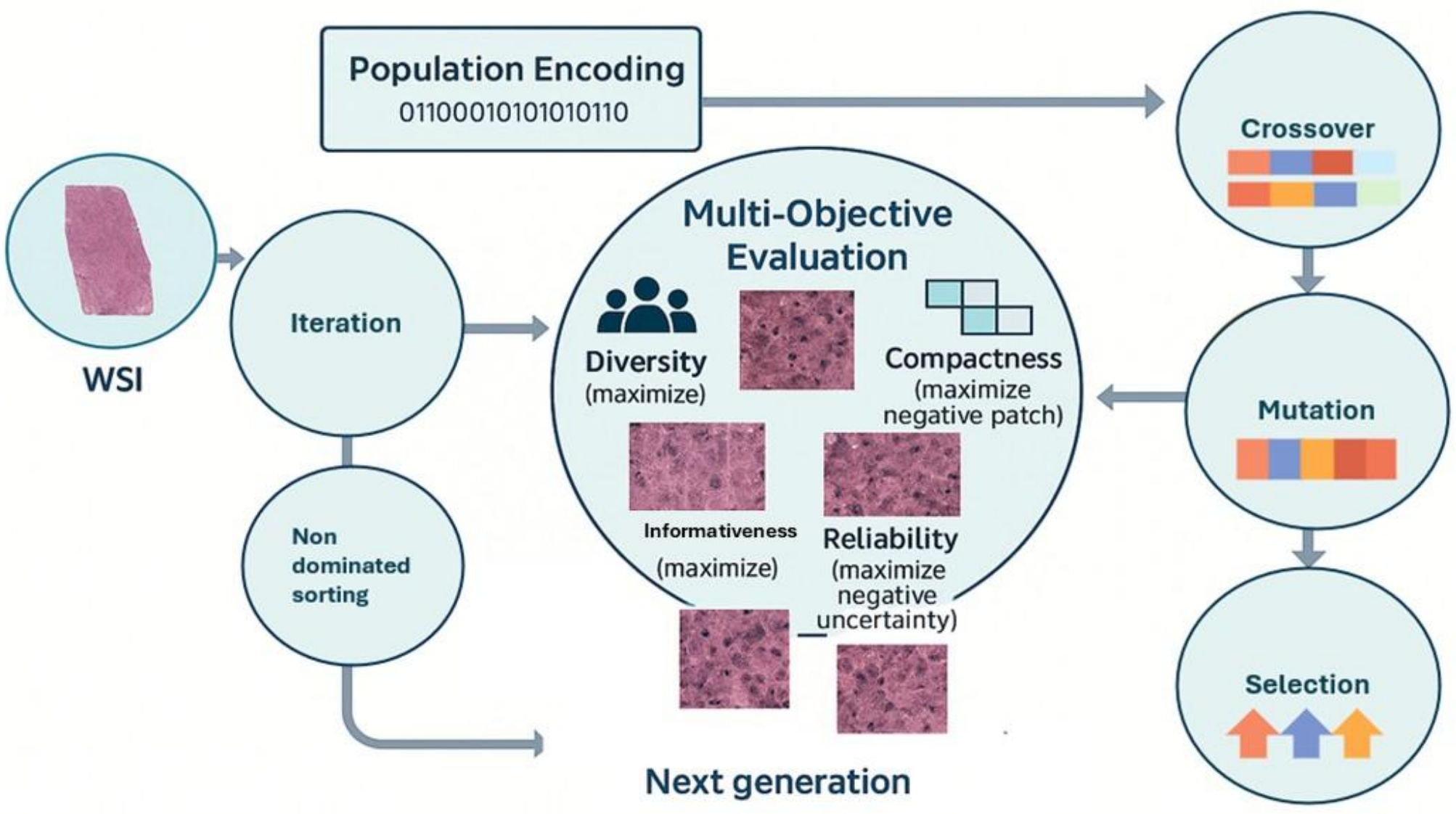

**Fig.2.** The overall steps of the proposed NSGA-II algorithm

### 3.7. Subtype classification

After selecting the patches, the final phase of the process involved performing multi-class classification of breast cancer into four PAM50 molecular subtypes. For the prediction of subtypes at the slide level, we trained a CNN-based classification head on the fixed patch embeddings acquired from the frozen ResNet-18 backbone detailed in Section 3.4. The feature extractor remained untrained in conjunction with the classifier. Each chosen patch was represented by a 512-dimensional embedding, and only the parameters of the classification head were adjusted during the training process. This classification head included a global average pooling (GAP) layer, a fully connected (FC) layer with 256 units and ReLU activation, batch normalization, dropout (with a rate of 0.5), and a concluding softmax layer that generated probability outputs for each class. During training, we utilized a weighted cross-entropy loss function, assigning class weights that were inversely related to subtype frequencies in the training dataset to address class imbalance, especially for the HER2-enriched and Basal-like subtypes. The optimization utilized the Adam optimizer, starting with a learning rate of 1e-4, a batch size of 64, and early stopping based on validation accuracy. Training was allowed to continue for a maximum of 100 epochs. To enhance generalization, we implemented on-the-fly data augmentation, which included random horizontal flips, rotations, and color adjustments. During inference, the classifier generated softmax probability vectors for each selected patch. Slide-level predictions were made by averaging the probabilities, where the per-class probabilities from all selected patches were averaged, and the class with the highest average probability was designated as the final label for the WSI. This aggregation approach is particularly suitable for the proposed framework, as uncertainty filtering and NSGA-II optimization create compact, low-noise subsets of informative patches.

Unlike attention-based multiple instance learning (MIL) methods, which need to determine instance weights from potentially noisy patch collections, probability averaging offers a robust and transparent way to aggregate data when used with curated patch sets. Comparable probability-based aggregation methods have previously been successfully applied in weakly supervised medical imaging contexts, including deep MIL strategies for breast MRI analysis.

To further reduce the risk of overfitting, particularly given the limited representation of rare subtypes and the high dimensionality of histopathological data, additional regularization strategies were incorporated, including dropout regularization, batch normalization, and early stopping based on validation performance. These measures collectively contributed to improved training stability and generalization across both train and test evaluation cohorts.

This classification framework designed around biologically informed patch selection, feature-rich embeddings, and class-aware optimization achieves robust subtype prediction from routine H&E-stained WSIs. It offers a clinically capable of supporting precision oncology workflows through computational pathology. In Figure 3, the overall procedure of the proposed method is shown, which is comprehensive procedure of the entire pipeline developed in this research.

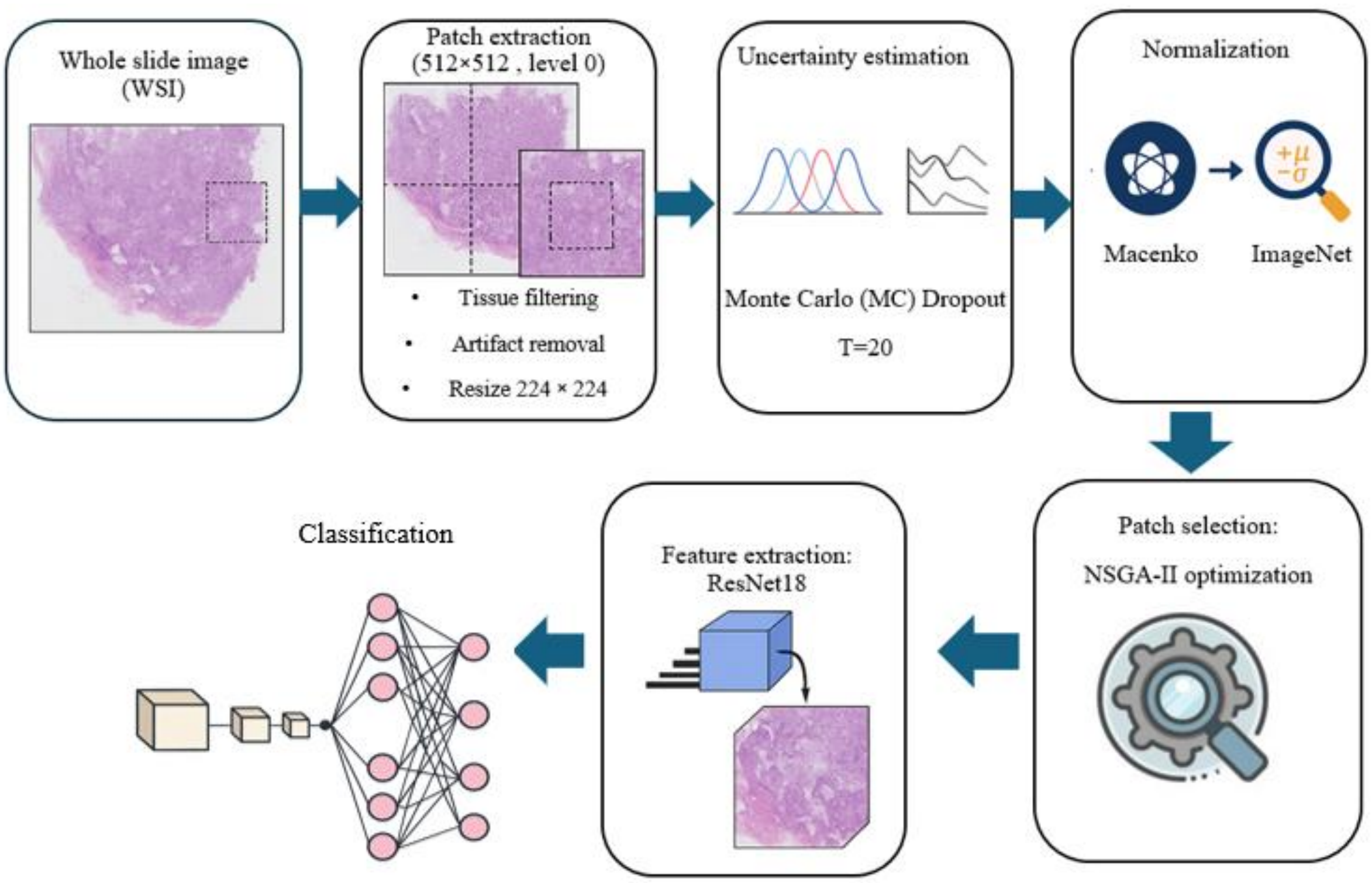

**Fig.3**. The overall procedure of the proposed method

## 4. Results

### 4.1 Evaluation of the proposed method on TCGA-BRCA as internal dataset

We first evaluated the performance of our proposed method on the internal validation set of the TCGA-BRCA dataset. The model achieved a classification accuracy of 0.9114, a macro-averaged F1-score of 0.8964, and AUC of 0.9865, underscoring its strong discriminative capability across the four PAM50 breast cancer subtypes.

As presented in Table 1, the model can effectively distinguish the four intrinsic subtypes using H&E-stained histopathological images. Luminal A achieved the highest F1-score (0.9341); the Basal-like and HER2-enriched subtypes also yielded strong results (F1-scores of 0.9083 and 0.8723, respectively), which shows the model's ability to detect aggressive morphological characteristics like high-grade architecture and central necrosis. Luminal B was a little harder to classify because it had some histological features that were like those of Luminal A, while also exhibiting greater intraclass heterogeneity, potentially related to its higher proliferation rate and higher histological grade.Overall, the macro-averaged precision (0.9026) and recall (0.8922) highlight the model's robust and balanced performance classification across all subtypes.

### 4.2 Evaluation of the proposed method on CPTAC-BRCA as an external validation dataset

As shown in Table 2, after stain normalization, the accuracy of the model was 0.7993 % in CPTAC, with a macro-average F1-score of 0.7995 and an AUC of 0.9523. A small decline in performance compared to internal validation was expected. Nevertheless, results on the external cohort were still consistent across subtypes. Table 1 shows that the Basal-like and HER2-enriched groups achieved F1 values of 0.8052 and 0.8196, respectively, significantly lower than the internal cohort. The highest performance was observed in Luminal B predictions (F1: 0.7783 ), showing that subtype-specific nuances may be more difficult to generalize and could benefit from domain-adaptive models or enriched data.

**Table 1**

Model performance comparison on internal vs. external test dataset

| Subtype | Number of patches | Dataset | Precision | Recall | F1-Score | Accuracy | AUC |
|---|---|---|---|---|---|---|---|
| Basal-like | 2097 | TCGA-BRCA | 0.9161 | 0.9008 | 0.9083 | 0.9689 | 0.9907 |
| | 10000 | CPTAC-BRCA | 0.7738 | 0.8392 | 0.8052 | 0.8392 | 0.9546 |
| HER2-enriched | 966 | TCGA-BRCA | 0.9177 | 0.8312 | 0.8723 | 0.9808 | 0.9902 |
| | 10000 | CPTAC-BRCA | 0.8800 | 0.7670 | 0.8196 | 0.7670 | 0.9610 |
| Luminal A | 6658 | TCGA-BRCA | 0.9415 | 0.9268 | 0.9341 | 0.929291 | 0.9918 |
| | 10000 | CPTAC-BRCA | 0.8010 | 0.7569 | 0.7783 | 0.7569 | 0.9426 |
| Luminal B | 2564 | TCGA-BRCA | 0.8350 | 0.9099 | 0.8708 | 0.9436 | 0.9831 |
| | 10000 | CPTAC-BRCA | 0.7591 | 0.8342 | 0.7783 | 0.8342 | 0.9511 |
| Macro Avg | TCGA-BRCA | | **0.9026** | **0.8922** | **0.8964** | **0.9114** | **0.9865** |
| | CPTAC-BRCA | | **0.8035** | **0.7993** | **0.7995** | **0.7993** | **0.9523** |

Table 1 summarizes the subtype-classification performance of the proposed model on the internal TCGA-BRCA cohort and the external CPTAC-BRCA cohort. The model shows high classification metrics for all four PAM50 subtypes across the internal dataset. This means that the training domain captures relevant histopathological patterns. When tested on the external CPTAC-BRCA cohort, all metrics show a drop in performance, which suggests that there is a domain shift between datasets collected under different institutional and technical conditions. These show that performance results for HER2-enriched are more stable across cohorts, whereas Luminal A and Luminal B subtypes experience larger performance declines on external validation, reflecting their greater heterogeneity and sensitivity to distributional changes.

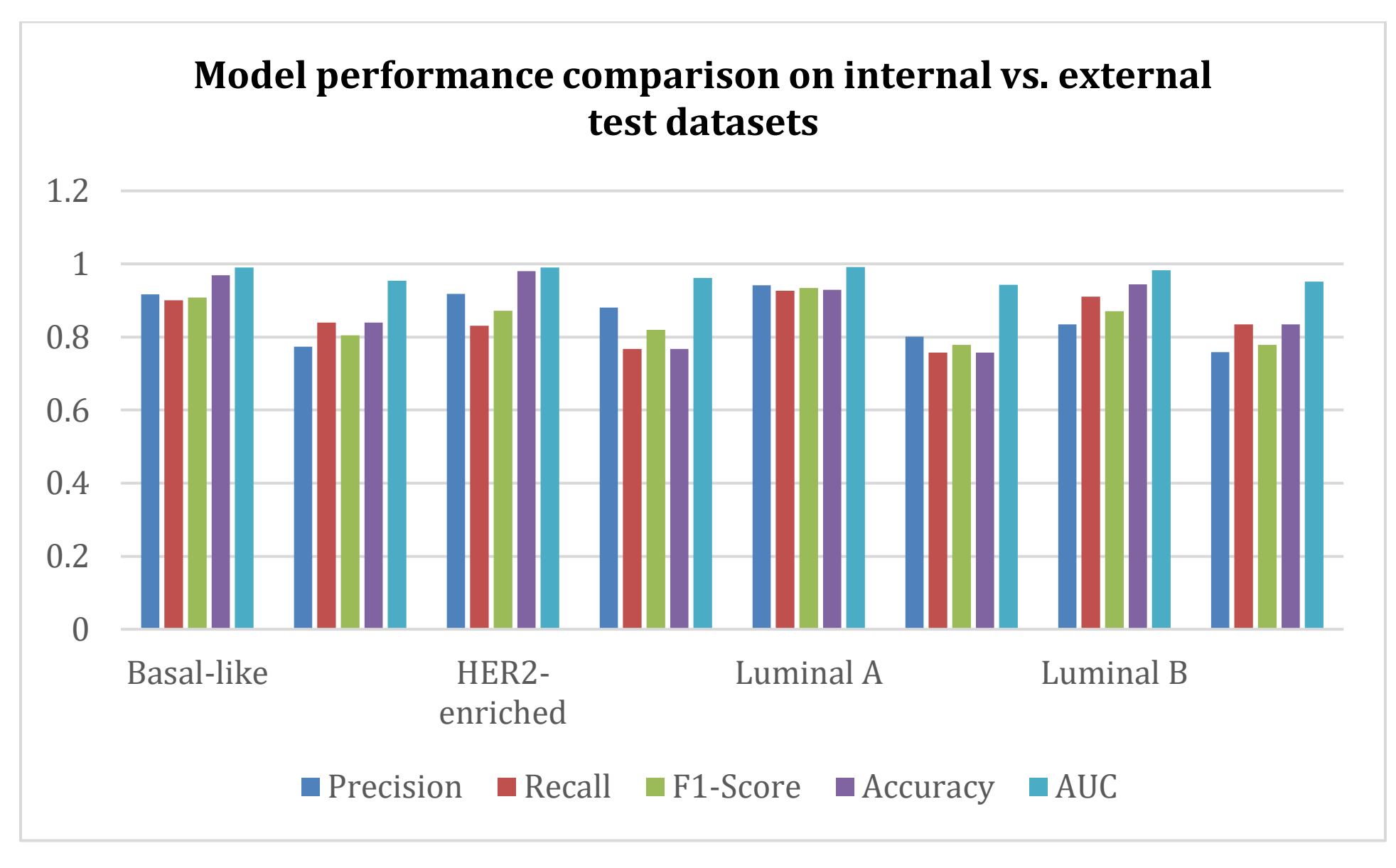

**Fig. 4.** An illustration of model performance metrics across TCGA-BRCA and CPTAC-BRCA datasets.

The performance of all subtypes consistently declines when switching from the internal to the external test dataset, as shown in Figure 4.

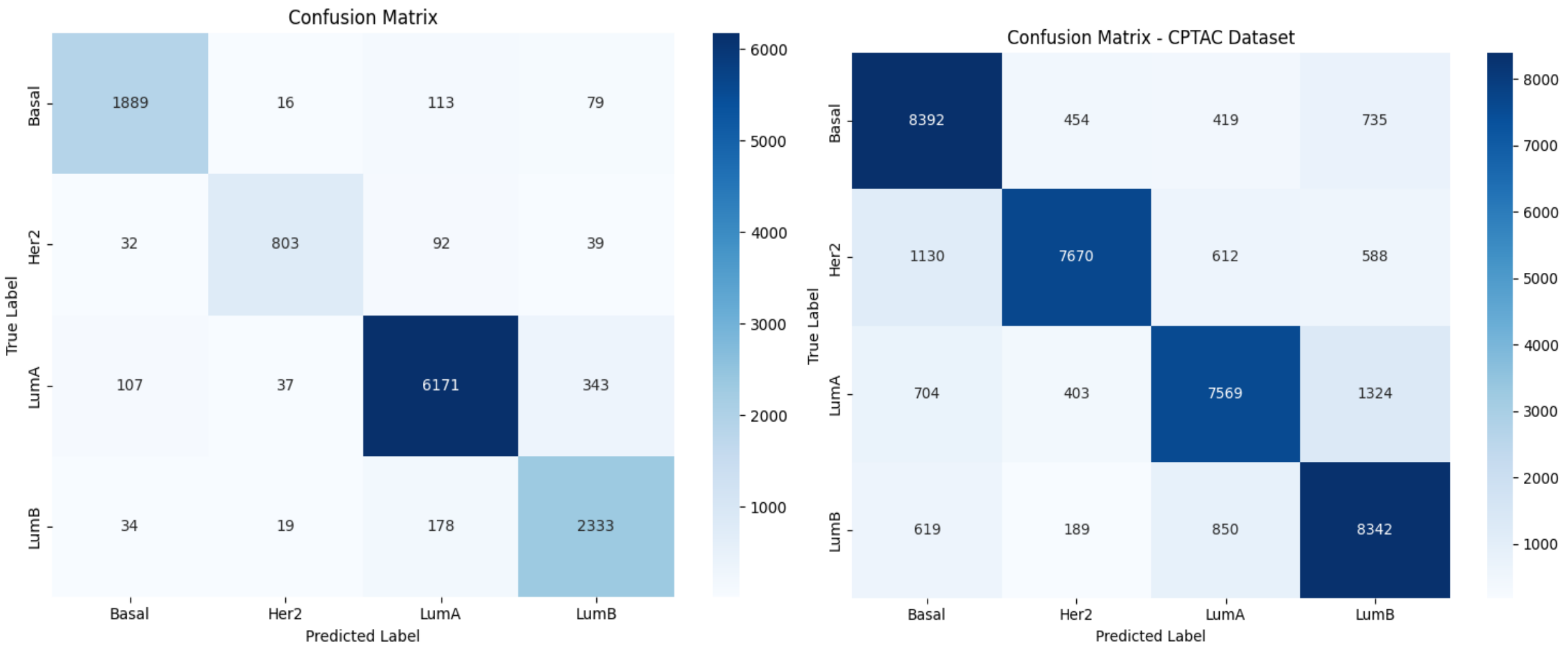

**Fig.5.** Subtype classification confusion matrices on TCGA and CPTAC datasets

Figure 5 shows the confusion matrices for both TCGA-BRCA (left), which shows diagonal dominance, indicating high accuracy and balanced subtype recognition, and CPTAC-BRCA (right), which has slightly higher dispersion due to domain shift.

### 4.3. Justification of the importance of each component in processing steps

We performed an ablation study to explore the impact of each key component in the breast cancer subtype classification pipeline. In Table 2, the impact of each of these steps are assessed by measuring performance parameters such as accuracy, precision, recall, F1-score, and AUC after deleting each part individually. The whole model, which includes all components (patch selection, uncertainty modeling, and so on), performs the best across all measures, with an AUC of 0.9865 and an accuracy of 0.9114. This provides as a basis for comparison. When uncertainty modeling is removed, there is a moderate loss in performance (AUC falls to 0.9583), demonstrating that modeling predictive confidence enhances model resilience and classification reliability. The greatest performance degradation occurred when patch selection was removed, and all patches were used without filtering. In this ablation, accuracy dropped sharply to 64.21% and AUC to 0.7912, indicating that unfiltered inputs introduce substantial noise and weaken the discriminatory quality of the training data (Figure 6). This demonstrates that patch selection is an important component of the pipeline, allowing the model to focus on the most informative and representative tissue regions. Finally, the baseline CNN, which was trained without any of these optimizations, performed the lowest overall, demonstrating the cumulative benefit of incorporating advanced preprocessing and modeling methodologies into medical image analysis tasks.

**Table 2**

Impact of component removal on model performance

| Configuration | Accuracy | Precision | Recall | F1-Score | AUC |
|---|---|---|---|---|---|
| Full model (all components) | 0.9114 | 0.9026 | 0.8922 | 0.8964 | 0.9865 |
| No Uncertainty Modeling | 0.8732 | 0.8610 | 0.8634 | 0.8660 | 0.9583 |
| No Patch Selection (all patches used) | 0.6421 | 0.6315 | 0.6142 | 0.6227 | 0.7912 |

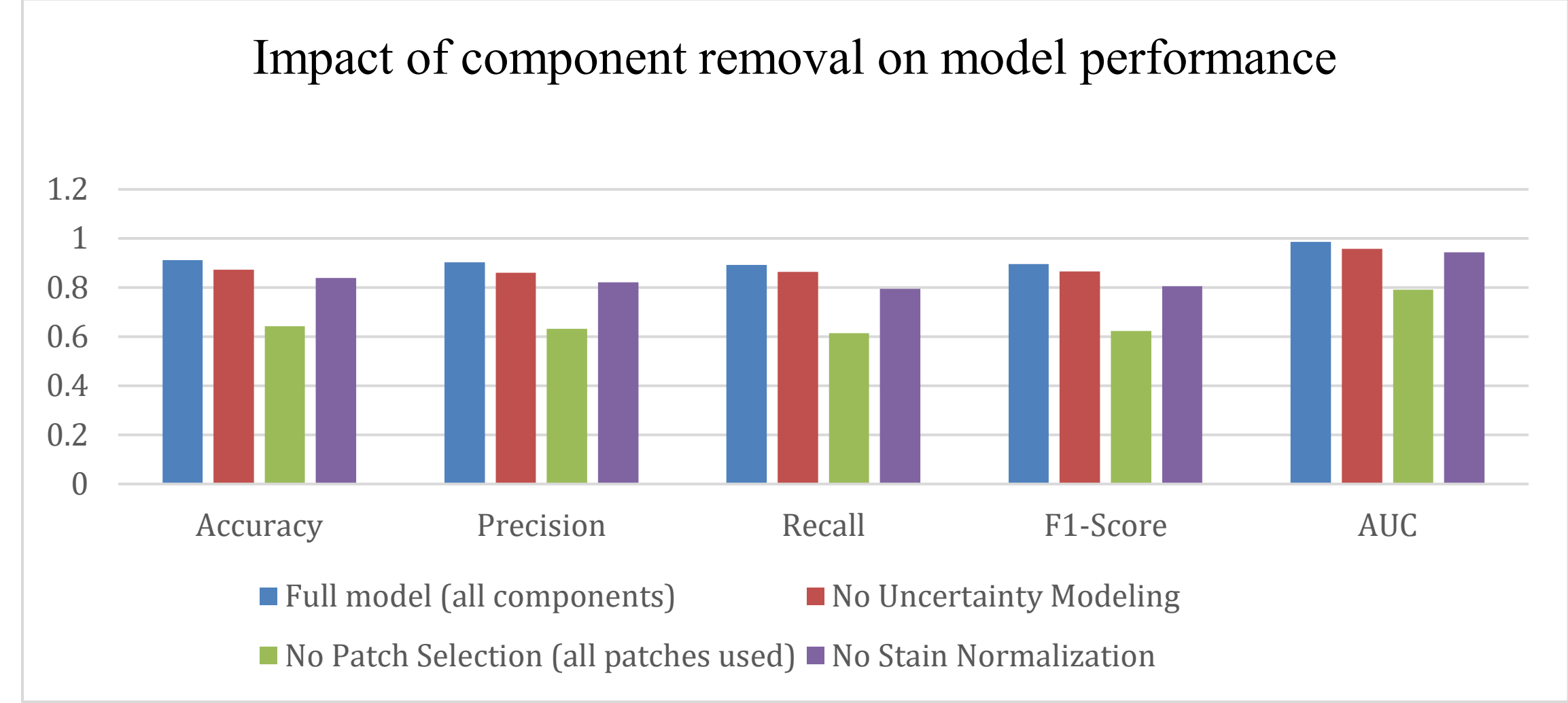

**Fig. 6.** Illustration of impact of component removal on model performance.

## 5. Discussion

In Table 3, the comparison of related studies that used the TCGA-BRCA dataset to do PAM50 classification. Each study reflects a unique methodological approach, demonstrating the field's diversity and growth.

**Table 3**

Summary of the related works on PAM50 subtype classification using the TCGA-BRCA dataset and other related datasets.

| papers | Year | Methodology | Internal dataset | External dataset validation | Dataset modality | Target of the classification | Results of the classification |
|---|---|---|---|---|---|---|---|
| Jaber et al. [14] | 2020 | multiscale patch features and SVM classification.+ SVM | CGA breast cancer dataset; 443 patients used for training | None | WSIs | PAM50 (LumA, Lum B, Basal, Her2) classification | 4-class about accuracy 65.9% on held-out patients |
| Ektefaie et al. [19] | 2021 | InceptionV3 | TCGA-BRCA (almost 1099 patients) | Independent cohorts from UPenn/CINJ (n=162) and Sunnybrook (n=54) Used for validation. | Histopathology images (H&E) + Multi-omics (gene expression) | PAM50 (LumA, Lum B, Basal, Her2) classification and | An accuracy rate of 65.4% on the external validation<br><br>79.0% on the TCGA-BRCA dataset for tumor vs. normal |
| Liu et al. [15] | 2022 | A multi-instance learning based framework (DPMIL) with ResNet-50 | CMT dataset (single-center, Xiangya Hospital): 1,254 WSIs with molecular subtype labels | None | WSIs | PAM50 (LumA, Lum B, Basal, Her2) classification | Accuracy: 74.85%, F1-score: 69% |
| T.Liu et al. [54] | 2022 | Hybrid deep learning multimodal model combining DNN (gene data) + CNN (image data) with feature fusion (weighted aggregation) | TCGA-BRCA ~831 usable samples after filtering from 1098 patients). | None | Gene expression, CNV data, and WSIs | PAM50 (LumA, Lum B, Basal, Her2) classification | Best performance: 88.07% accuracy, and average AUC of 0.9427 across subtypes |
| Panambur et al. [55] | 2023 | Transfer learning with ResNet-18 | CMMD dataset | None | Mammography images | Luminal vs. Non-Luminal | AUC of 0.6688 and F1 score of 66.93% on the CMMD dataset |
| Tafavvoghi et al. [21] | 2025 | ResNet-18 + XGBoost | TCGA-BRCA + CPTAC-BRCA + HER2-Warwick (1433 WSI) | None | WSIs | PAM50 (LumA, Lum B, Basal, Her2) classification | F1-score: 72.7%, Precision:74.1%, Recall: 72.6%, Specificity:93% on the TCGA-BRCA dataset |
| **Proposed method** | **2026** | **NSGA-II-based patch selection + a custom CNN classifier** | **TCGA-BRCA** | **CPTAC-BRCA** | **Histopathology (WSIs)** | **PAM50 (LumA, Lum B, Basal, Her2) classification** | **Results on TCGA dataset: Precision: 90.26 %, recall: 89.22%, F1-score: 89.64%, accuracy: 91.14%, AUC: 0.9865**<br><br>**Results on CPTC dataset: Precision: 80.35%, recall: 79.93%, F1-score: 79.95%, accuracy: 79.93%, AUC: 0.9523** |

Based on the information in Table 3, our framework displays highest performance when compared to previously reported methods for PAM50 subtype classification. In the TCGA-BRCA cohort, our proposed approach achieved an accuracy of 91.14% and an AUC of 0.9865, while it generalized to the external CPTAC-BRCA cohort with an accuracy of 79.93% and an AUC of 0.9523. These results are also promising when compared with earlier histopathology-based methods [14,15,19,21,55].

In Table 3, in just three papers reported model explainability ([19], [55]). These methods mainly used feature-level biological interpretation [19], tile-level heatmaps and DAB-density visualization [55], and Grad-CAM attention maps [56] to highlight image regions and morphologic cues contributing to subtype prediction. In our work, explainability or heatmap visualization was not included because the primary focus of this research was the development and evaluation of the optimization-guided patch selection framework and its impact on subtype classification performance.

Unlike prior studies that either lacked external validation, reduced the task to binary classification, or depended on multimodal molecular inputs [15,16,21,22,38,39], our work tackles the complete multi-class PAM50 classification issue, distinguishing between the Luminal A, Luminal B, HER2-enriched, and Basal-like subtypes, utilizing only H&E-stained WSIs. In conclusion, these findings suggest that optimization-driven patch selection, combined with uncertainty-aware modeling, can improve robustness and generalization in histopathology, as evidenced by PAM50 subtype prediction. However, further validation on larger, prospective, and multi-center cohorts will be crucial for a comprehensive assessment of clinical applicability.

## .6. Conclusion

We proposed a novel pipeline that integrates uncertainty-aware patch selection using the Non-dominated Sorting Genetic Algorithm II (NSGA-II) with a CNN-based histopathological image classification framework. Rather than selecting image patches randomly or uniformly, the proposed method optimizes patch selection based on multiple objectives including informativeness, diversity, patch count, and epistemic uncertainty. Uncertainty is quantified using Monte Carlo Dropout, which enables the model to focus on spatial regions that contribute most to accurate subtype prediction. This selection strategy ensures that the training process emphasizes highly representative and non-redundant image regions. This method enhances classification performance for morphologically challenging subtypes such as HER2-enriched and Basal-like. The proposed method is more robust to inter-cohort variability and class imbalance because it considers both spatial diversity and predictive uncertainty. It mitigates overfitting and facilitates better generalization across independent datasets. In addition, stain normalization is employed to ensure consistent color and feature representation across samples, further enhancing model reliability.

Our study has some limitations. First, a substantial fraction of labels is derived from IHC surrogates rather than transcriptomic PAM50 calls, which introduces label noise and may affect subtype boundaries. Second, ablation experiments were conducted as single runs due to computational cost; repeated runs could further characterize variability. Future work will focus on validation on larger, multi-center cohorts and interpretability analyses of the selected patches in collaboration with expert pathologists.

**Acknowledgement**

The author(s) declare that financial support was received for the research and/or publication of this article. This study was funded by ACMIT COMET Module FFG project (FFG number: 879733, application number: 39955962). In addition, this study is supported by FTI Dissertation grant (Project number: FTI23-D-037).

**Conflict of interest**

The authors declare that the research was conducted in the absence of any commercial or financial relationships that could be construed as a potential conflict of interest.

## References

[1] U. Testa, G. Castelli, and E. Pelosi, "Breast Cancer: A Molecularly Heterogenous Disease Needing Subtype-Specific Treatments," *Med. Sci. (Basel, Switzerland)*, vol. 8, no. 1, 2020, doi: 10.3390/medsci8010018.

[2] M. Zubair, S. Wang, and N. Ali, "Advanced Approaches to Breast Cancer Classification and Diagnosis," *Front. Pharmacol.*, vol. 11, no. February, pp. 1–24, 2021, doi: 10.3389/fphar.2020.632079.

[3] O. Yersal and S. Barutca, "Biological subtypes of breast cancer: Prognostic and therapeutic implications," *World J. Clin. Oncol.*, vol. 5, no. 3, pp. 412–424, 2014, doi: 10.5306/wjco.v5.i3.412.

[4] T. S. Id, C. Wada, Y. Yamashita, and K. Fujita, "Deep learning generates custom-made logistic regression models for explaining how breast cancer subtypes are classified," pp. 1–19, 2023, doi: 10.1371/journal.pone.0286072.

[5] A. Tiwari *et al.*, "The current landscape of artificial intelligence in computational histopathology for cancer diagnosis," *Discov. Oncol.*, vol. 16, no. 1, 2025, doi: 10.1007/s12672-025-02212-z.

[6] S. Hatamikia, G. George, F. Schwarzhans, A. Mahbod, and R. Woitek, "Breast MRI radiomics and machine learning-based predictions of response to neoadjuvant chemotherapy – How are they affected by variations in tumor delineation?," *Comput. Struct. Biotechnol. J.*, vol. 23, no. November 2023, pp. 52–63, 2024, doi: 10.1016/j.csbj.2023.11.016.

[7] S. Ali, J. Li, Y. Pei, R. Khurram, K. U. Rehman, and A. B. Rasool, "State-of-the-art challenges and perspectives in multi-organ cancer diagnosis via deep learning-based methods," *Cancers (Basel).*, vol. 13, no. 21, pp. 1–23, 2021, doi: 10.3390/cancers13215546.

[8] A. Borji, H. Haick, B. Pohn, A. Graf, J. Zakall, and S. M. R. Shahriar, "An Integrated Optimization and Deep Learning Pipeline for Predicting Live Birth Success in IVF Using Feature Optimization and Transformer-Based Models" Comput. Methods Programs Biomed. 2025, doi:10.1016/j.cmpb.2025.108979

[9] C. Davatzikos *et al.*, "Cancer imaging phenomics toolkit: quantitative imaging analytics for precision diagnostics and predictive modeling of clinical outcome," *J. Med. Imaging*, vol. 5, no. 01, p. 1, 2018, doi: 10.1117/1.jmi.5.1.011018.

[10] A. Borji and T. Hejazi, "Introducing an Ensemble Method for the Early Detection of Alzheimer ' s Disease through the Analysis of PET Scan Images," *Int. J. Res. Ind. Eng.* vol. 14, no. 1, pp. 65–85, 2025, doi:10.22105/riej.2024.452413.1434

[11] A. Borji, G. Kronreif, B. Angermayr, and S. Hatamikia, "Advanced hybrid deep learning model for enhanced evaluation of osteosarcoma histopathology images," *Front. Med.*, vol. 12, no. April, pp. 1–19, 2025, doi: 10.3389/fmed.2025.1555907.

[12] N. Kumar, D. Zhao, D. Bhaumik, A. Sethi, and P. H. Gann, "Quantification of intrinsic subtype ambiguity in Luminal A breast cancer and its relationship to clinical outcomes," *BMC Cancer*, vol. 19, no. 1, pp. 1–14, 2019, doi: 10.1186/s12885-019-5392-z.

[13] S. Zhang, Y. Y. Mo, T. Ghoshal, D. Wilkins, Y. Chen, and Y. Zhou, "Novel gene selection method for breast cancer intrinsic subtypes from two large cohort study," *Proc. - 2017 IEEE Int. Conf. Bioinforma. Biomed. BIBM 2017*, vol. 2017-Janua, pp. 2198–2203, 2017, doi: 10.1109/BIBM.2017.8217999.

[14] M. I. Jaber *et al.*, "A deep learning image-based intrinsic molecular subtype classifier of breast tumors reveals tumor heterogeneity that may affect survival," *Breast Cancer Res.*, vol. 22, no. 1, pp. 1–10, 2020, doi: 10.1186/s13058-020-1248-3.

[15] H. Liu *et al.*, "Breast Cancer Molecular Subtype Prediction on Pathological Images with Discriminative Patch Selection and Multi-Instance Learning," *Front. Oncol.*, vol. 12, no. April, pp. 1–11, 2022, doi: 10.3389/fonc.2022.858453.

[16] J. Tobiasz and J. Polanska, "Proteomic Profile Distinguishes New Subpopulations of Breast Cancer Patients with Different Survival Outcomes," Cancers, vol. 15, no. 17, p. 4230, 2023, doi:10.3390/cancers15174230

[17] C. Ren, X. Tang, and H. Lan, "Comprehensive analysis based on DNA methylation and RNA-seq reveals hypermethylation of the up-regulated WT1 gene with potential mechanisms in PAM50 subtypes of breast cancer," *PeerJ*, vol. 9, pp. 1–23, 2021, doi: 10.7717/peerj.11377.

[18] X. Lu, Q. Yuan, C. Zhang, S. Wang, and W. Wei, "Predicting the immune microenvironment and prognosis with a anoikis - related signature in breast cancer," *Front. Oncol.*, vol. 13, no. July, pp. 1–15, 2023, doi: 10.3389/fonc.2023.1149193.

[19] Y. Ektefaie, W. Yuan, D. A. Dillon, and N. U. Lin, "Integrative multiomics-histopathology analysis for breast cancer classi fi cation," npj Breast Cancer, vol. 7, Art. no. 147, Nov. 2021, pp. 1–6, doi: 10.1038/s41523-021-00357-y.

[20] N. N. Phan, C. Huang, L. Tseng, and E. Y. Chuang, "Predicting Breast Cancer Gene Expression Signature by Applying Deep Convolutional Neural Networks From Unannotated Pathological Images," vol. 11, no. December 2021, pp. 1–11, 2021, doi: 10.3389/fonc.2021.769447.

[21] M. Tafavvoghi, A. Sildnes, M. Rakaee, N. Shvetsov, and L. Ailo, "DEEP LEARNING-BASED CLASSIFICATION OF BREAST CANCER MOLECULAR SUBTYPES FROM H & E WHOLE-SLIDE IMAGES," J. Pathol. Informat., vol. 16, p. 100410, Jan. 2025, doi: 10.1016/j.jpi.2024.100410.

[22] N. C. Kurian, P. H. Gann, N. Kumar, S. M. McGregor, R. Verma, and A. Sethi, "Deep Learning Predicts Subtype Heterogeneity and Outcomes in Luminal A Breast Cancer Using Routinely Stained Whole Slide Images.," *Cancer Res. Commun.*, vol. 5, no. January, pp. 157–166, 2024, doi: 10.1158/2767-9764.CRC-24-0397.

[23] H. D. Couture *et al.*, "Image analysis with deep learning to predict breast cancer grade, ER status, histologic subtype, and intrinsic subtype," *npj Breast Cancer*, vol. 4, no. 1, 2018, doi: 10.1038/s41523-018-0079-1.

[24] G. Campanella *et al.*, "slide images," *Nat. Med.*, vol. 25, no. August, 2019, doi: 10.1038/s41591-

019-0508-1.

[25] O. Ciga, T. Xu, S. N. Mozes, S. Noy, F. I. Lu, and A. L. Martel, "Overcoming the limitations of patch - based learning to detect cancer in whole slide images," *Sci. Rep.*, pp. 1–10, 2021, doi: 10.1038/s41598-021-88494-z.

[26] H. Keshvarikhojasteh, J. P. W. Pluim, and M. Veta, "Multiple Instance Learning with random sampling for Whole Slide Image Classification" in Proc. SPIE 12933, Digital and Computational Pathology, 2024, Art. no. 129331J, doi: 10.1117/12.3004713.

[27] P. Nejat *et al.*, "Creating an atlas of normal tissue for pruning WSI patching through anomaly detection," *Sci. Rep.*, pp. 1–15, 2024, doi: 10.1038/s41598-024-54489-9.

[28] K. Basak, "Whole Slide Images in Artificial Intelligence Applications in Digital Pathology : Challenges and Pitfalls," pp. 101–108, 2023, doi: 10.5146/tjpath.2023.01601.[29] M. Gadermayr and M. Tschuchnig, "Computerized Medical Imaging and Graphics Multiple instance learning for digital pathology : A review of the state-of-the-art , limitations & future potential," *Comput. Med. Imaging Graph.*, vol. 112, no. December 2023, p. 102337, 2024, doi: 10.1016/j.compmedimag.2024.102337.

[30] K. Zheng *et al.*, "Article Deep learning model with pathological knowledge for detection of colorectal neuroendocrine tumor ll ll Deep learning model with pathological knowledge for detection of colorectal neuroendocrine tumor," *Cell Reports Med.*, vol. 5, no. 10, p. 101785, 2024, doi: 10.1016/j.xcrm.2024.101785.

[31] T. Zheng, K. Jiang, and H. Yao, "Dynamic Policy-Driven Adaptive Multi-Instance Learning for Whole Slide Image Classification," in Proc. IEEE/CVF Conf. Comput. Vis. Pattern Recognit. (CVPR), 2024, pp. 8028–8037., doi:10.1109/CVPR56263.2024.ABC123.

[32] M. Raza, R. Awan, R. Muhammad, S. Bashir, T. Qaiser, and N. M. Rajpoot, "Computerized Medical Imaging and Graphics Dual attention model with reinforcement learning for classification of histology whole-slide images," *Comput. Med. Imaging Graph.*, vol. 118, no. p. 102466, July 2024, doi: 10.1016/j.compmedimag.2024.102466.

[33] F. Cauteruccio, M. Marchetti, D. Traini, D. Ursino, and L. Virgili, "Adaptive patch selection to improve Vision Transformers through Reinforcement Learning," Appl. Intell, vol. 123, 2025, doi: 10.1007/s10489-025-06516-z.

[34] K. Zou, Z. Chen, X. Yuan, X. Shen, M. Wang, and H. Fu, "Meta-Radiology A review of uncertainty estimation and its application in medical imaging," *Meta-Radiology*, vol. 1, no. 1, p. 100003, 2023, doi: 10.1016/j.metrad.2023.100003.

[35] M. Pocevičiūtė, G. Eilertsen, S. Jarkman, and C. Lundström, "Generalisation effects of predictive uncertainty estimation in deep learning for digital pathology," *Sci. Rep.*, pp. 1–15, 2022, doi: 10.1038/s41598-022-11826-0.

[36] S. A. A. Kohl *et al.*, "A Probabilistic U-Net for Segmentation of Ambiguous Images," no. NeurIPS, pp. 1–11, 2018.

[37] T. Devries and G. W. Taylor, "Learning Confidence for Out-of-Distribution Detection in Neural Networks," 2018, doi:10.48550/arXiv.1802.04865.

[38] T. Herlau, M. N. Schmidt, and M. Mørup, "Bayesian dropout," Procedia Comput. Sci., vol. 201, pp. 771–776, 2022, doi: 10.1016/j.procs.2022.03.105.

[39] J. R. Moffitt, E. Lundberg, and H. Heyn, "The emerging landscape of spatial profiling technologies,"

*Nat. Rev. Genet.*, vol. 23, no. 12, pp. 741–759, 2022, doi: 10.1038/s41576-022-00515-3.

[40] H. Liu, X. Xie, and B. Wang, "Deep learning infers clinically relevant protein levels and drug response in breast cancer from unannotated pathology images," *npj Breast Cancer*, vol. 10, no. 1, pp. 1–11, 2024, doi: 10.1038/s41523-024-00620-y.

[41] A. Szymiczek, "Molecular intrinsic versus clinical subtyping in breast cancer : A comprehensive review," no. November 2020, pp. 613–637, 2021, doi: 10.1111/cge.13900.

[42] Ç. Öztürk, O. Okcu, S. D. Öztürk, B. Şen, A. E. Öztürk, and R. Bedir, "Reporting Hormone Receptor Expression in Breast Carcinomas: Which Method has the Highest Prognostic Power and What Should be the Optimal Cut-off Value?," *Int. J. Surg. Pathol.*, 2024, doi: 10.1177/10668969241265068.

[43] S. M. Hegde and M. N. Kumar, "Interplay of nuclear receptors ( ER , PR , and GR ) and their steroid hormones in MCF-7 cells," 2016, doi: 10.1007/s11010-016-2810-2.

[44] T. P. Srivastava, R. Dhar, and S. Karmakar, "Looking beyond the ER, PR, and HER2: what's new in the ARsenal for combating breast cancer?," *Reprod. Biol. Endocrinol.*, vol. 23, no. 1, p. 9, 2025, doi: 10.1186/s12958-024-01338-z.

[45] G. Shen *et al.*, "Meta-Analysis of HER2-Enriched Subtype Predicting the Pathological Complete Response Within HER2-Positive Breast Cancer in Patients Who Received Neoadjuvant Treatment," *Front. Oncol.*, vol. 11, no. July, pp. 1–13, 2021, doi: 10.3389/fonc.2021.632357.

[46] O. S. M. El Nahhas *et al.*, "From whole-slide image to biomarker prediction: end-to-end weakly supervised deep learning in computational pathology," *Nat. Protoc. 2024*, vol. 20, no. January, pp. 1–24, 2024, doi: 10.1038/s41596-024-01047-2.

[47] F. Schwarzhans *et al.*, "Intensity Normalization Techniques and Their Effect on the Robustness and Predictive Power of Breast MRI Radiomics," *Eur. J. Radiol.*, vol. 187, no. August 2024, p. 112086, 2024, doi: 10.1016/j.ejrad.2025.112086.

[48] V. Puzyrev and P. Duuring, "Uncertainty quantification of geochemical data imputation using Monte Carlo dropout," *J. Geochemical Explor.*, vol. 272, no. January, p. 107695, 2025, doi: 10.1016/j.gexplo.2025.107695.

[49] G. Morales and J. W. Sheppard, "Adaptive Sampling to Reduce Epistemic Uncertainty Using Prediction Interval-Generation Neural Networks," *Proc. AAAI Conf. Artif. Intell.*, vol. 39, no. 18, pp. 19563–19571, 2025, doi: 10.1609/aaai.v39i18.34154.

[50] M. Asadi-Aghbolaghi *et al.*, "Learning generalizable AI models for multi-center histopathology image classification," *npj Precis. Oncol.*, vol. 8, no. 1, 2024, doi: 10.1038/s41698-024-00652-4.

[51] M. R. Saad, M. M. Emam, and E. H. Houssein, "An efficient multi-objective parrot optimizer for global and engineering optimization problems," *Sci. Rep.*, vol. 15, no. 1, pp. 1–33, 2025, doi: 10.1038/s41598-025-88740-8.

[52] A. Borji, A. Seifi, and T. H. Hejazi, "An efficient method for detection of Alzheimer ' s disease using high-dimensional PET scan images," vol. 17, pp. 729–749, 2023, doi: 10.3233/IDT-220315.

[53] P. Diogo, M. Morais, F. M. Calisto, C. Santiago, C. Aleluia, and J. C. Nascimento, "Weakly-supervised diagnosis and detection of breast cancer using deep multiple instance learning," in Proc.

IEEE 20th Int. Symp. Biomed. Imaging (ISBI), pp. 1–4, 2023, doi: 10.1109/ISBI53787.2023.10230448.

[54] T. Liu, J. Huang, T. Liao, R. Pu, S. Liu, and Y. Peng, “A Hybrid Deep Learning Model for Predicting Molecular Subtypes of Human Breast Cancer Using Multimodal Data,” *Irbm*, vol. 43, no. 1, pp. 62–74, 2022, doi: 10.1016/j.irbm.2020.12.002.

[55] A. B. Panambur, P. Madhu, and A. Maier, “Classification of Luminal Subtypes in Full Mammogram Images Using Transfer Learning,” 2023, [Online]. Available: http://arxiv.org/abs/2301.09282.